\def\year{2022}\relax
\documentclass[letterpaper]{article} 
\usepackage{aaai22}  
\usepackage{times}  
\usepackage{helvet}  
\usepackage{courier}  
\usepackage[hyphens]{url}  
\usepackage{graphicx} 
\usepackage{natbib}  
\usepackage{caption} 
\DeclareCaptionStyle{ruled}{labelfont=normalfont,labelsep=colon,strut=off} 
\usepackage{algorithm}
\usepackage{algorithmic}

\usepackage{newfloat}
\usepackage{listings}

\usepackage[dvipsnames]{xcolor}
\usepackage{wrapfig}
\usepackage{amsmath}
\usepackage{amsfonts}

\floatstyle{ruled}
\newfloat{listing}{tb}{lst}{}
\floatname{listing}{Listing}
\title{Meta-CPR: Generalize to Unseen Large Number of Agents with Communication Pattern Recognition Module}
\author{
    Wei-Cheng Tseng \textsuperscript{\rm 1},
    Wei Wei \textsuperscript{\rm 2},
    Da-Chen Juan \textsuperscript{\rm 2},
    Min Sun \textsuperscript{\rm 1,3}
}
\affiliations{
    \textsuperscript{\rm 1} National Tsing Hua University,\;
    \textsuperscript{\rm 2} Google AI Research,\;
    \textsuperscript{\rm 3} Appier Inc., Taiwan \\
    weichengtseng@gapp.nthu.edu.tw, \{wewei,dacheng\}@google.com,  sunmin@ee.nthu.edu.tw 
    
}

\usepackage{bibentry}

\begin{document}

\maketitle

\definecolor{incomplete}{HTML}{000000}
\definecolor{modified}{RGB}{0,0,0}
\definecolor{neurips}{RGB}{0,0,0}
\definecolor{aaai}{RGB}{0,0,0}
\definecolor{DC}{HTML}{70AD47}
\definecolor{WW}{HTML}{ED7D31}
\definecolor{MS}{HTML}{7030A0}

\begin{abstract}
	Designing an effective communication mechanism among agents in reinforcement learning has been a challenging task, especially for real-world applications. The number of agents can grow or an environment sometimes needs to interact with a changing number of agents in real-world scenarios. To this end, a multi-agent framework needs to handle various scenarios of agents, in terms of both scales and dynamics, for being practical to real-world applications. We formulate the multi-agent environment with a different number of agents as a multi-tasking problem and propose a meta reinforcement learning (meta-RL) framework to tackle this problem. The proposed framework employs a meta-learned Communication Pattern Recognition (CPR) module to identify communication behavior and extract information that facilitates the training process. Experimental results are poised to demonstrate that the proposed framework (a) generalizes to an unseen larger number of agents and (b) allows the number of agents to change between episodes. The ablation study is also provided to reason the proposed CPR design and show such design is effective.
\end{abstract}

\section{Introduction}
Multi-agent reinforcement learning (MARL) has been widely-used in many real-world applications, such as autonomous driving system \cite{eccv20_v2v,corl18_flow}, network system control \cite{iclr20_nsc}.
One challenge in MARL is to let agents perform reasonable cooperative behavior via communication. Particularly, in a partially-observable environment, each agent only gets to observe part of the environment; therefore, agents need to share information and learn from others' observations and make an effective communication. With an effective communication mechanism, agents can consistently perform a cooperative behavior depending on the rewards.
    

In real-world scenarios, the number of agents may not always be a predefined value, and may change from time to time \cite{eccv20_v2v}. When the number of agents increases, the communication becomes significantly more complex since the state space grows exponentially, which in turn leads to a large variation of policy gradients during the optimization process \cite{neurips17_maddpg}. This has been a fundamental challenge in the MARL paradigm, and needs to be resolved by a framework that generalizes to a large and unseen number of agents.
    
\begin{figure}
	\begin{center}
		\includegraphics[width=0.45\textwidth]{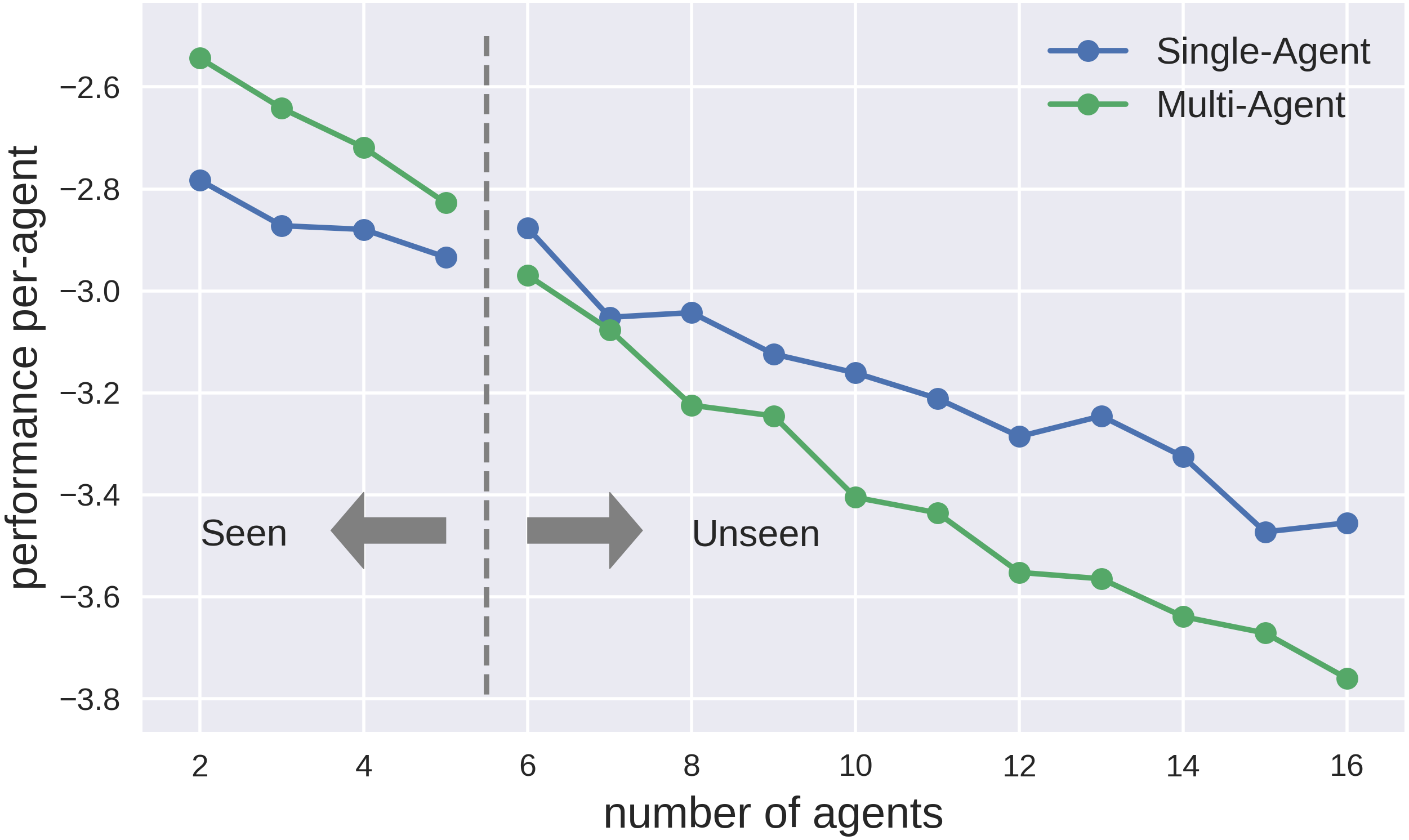}
	\end{center}
	\vspace{-6pt}
	\caption{
	\textcolor{neurips}{
        Compared to single-agent training (conventional RL method), a multi-agent RL framework performs better in an environment with a seen number of agent, but performs \textbf{worse} in an unseen large unseen number of agents scenario. 
	}
	\vspace{-24pt}
	}
	\label{fig:teaser}
\end{figure}

    
\textcolor{neurips}{
Multi-agent framework adapting to an unseen large number of agents hasn't been well-explored. The naive approach to tackle this problem is to train each agent independently without communication, and it is equivalent to the single-agent approach. We compare this approach with the off-the-shelf multi-agent method \cite{iclr19_ic3net} in Fig. \ref{fig:teaser}. Surprisingly, although single-agent method performs worse in an environment with a seen number of agents, it outperforms multi-agent framework for adapting to an unseen large number of agents.
This result implies the potential and necessity of the multi-agent framework which has the ability to efficiently adapt to an unseen large number of agents.
}

In this paper, we tackle the multi-agent system in the aspect of multi-task formulation by treating different number of agents as different tasks defined by Partially Observable Markov Decision Process (POMDP). Therefore, learning to generalize to different number of agents can be formulated as a multi-tasking problem. To further extend our framework to unseen numbers of agents, we leverage the techniques from meta learning. Specifically, we propose a Communication Pattern Recognition (CPR) module to meta-learn the information from communication histories, which facilitates the training of a policy network. Once the CPR module is learned, each agent can better utilize messages and the local observation with the instructions from CPR module. Experimental results show that the proposed method is able to generalize to an unseen and larger number of agents for both discrete and continuous action space, whereas other off-the-shelf multi-agent communication frameworks fall short in these scenarios. We further provide ablation studies to examine the design of the architecture of the CPR module and the training pipeline of the framework. Finally, the source codes and environment settings will be open sourced upon paper acceptance.
\section{Related Works}
Multi-task RL is attractive research to make a system acquire a variety of abilities with better sample efficiency. Meta-RL further models the task distribution to accelerate the learning speed based on the formulation of multi-task RL. Besides, multi-agent RL focus on making agents coordinate with each other in competitive or cooperative scenario.

\subsubsection{Multi-Task RL and Meta RL}
Multi-task learning is a challenging problem in RL. While training multiple tasks jointly allow the policies to share parameters across different tasks and potentially increase data efficiency \cite{iclr20_shared_knowledge}, the optimization problem becomes non-trivial \cite{aaai21_tseng}. Recently, some works propose to use modularization \cite{neurips20_soft_modularization} to solve increase sample efficiency. Other works focus on modifying the gradient to avoid gradient conflict \cite{neurips20_pcgrad} or unbalanced objectives \cite{icml18_gradnorm}.
    
Under the formulation of multi-tasking learning, meta-learn, or learn to learn, is a family of learning algorithms that intend to capture the learning procedure so that a model can quickly and efficiently learn another task if all the tasks share the same structure. One well-known kind of method is the context-based method \cite{iclr18_atten_meta}, which tries to utilize previous data to capture the task distribution. 
\textcolor{aaai}{
Besides, some works try to change the update rule of the training process \cite{icml20_metafun,icml17_maml} and further improve the adaptation efficiency to other tasks. 
}
    
Meta-learning can be extended to meta-RL. Meta-RL is well-formulated by recent works \cite{icml19_pearl,iclr20_metaq} under the formulation of multi-tasking RL. These methods try to extract the task information in the task distribution to make the framework generalize to different tasks or quickly adapt to similar tasks. Some other works intend to leverage meta-learn to learn the dynamic model \cite{iclr19_dynamic_adapt}, while some works integrate meta-learning with imitation learning to improve generalization \cite{iclr20_wtl}.

\subsubsection{Multi-Agent RL}
Applying RL to multi-agent scenarios has been an active research domain to encourage agents to perform consistently cooperative or competitive behavior for several years. 
\textcolor{aaai}{
To encourage cooperative behavior in the multi-agent scenario, one kind of method is to combine a joint reward signal and a credit assignment mechanism~\cite{aamas18_vdn,aamas18_lola,icml18_qmix,icml19_qtran,iclr21_updet} to value-based RL. COPA~\cite{icml21_copa} and REFIL~\cite{icml21_refil} use dynamic team composition to further benefit the training efficiency.
}
Another kind of methods try to extend the actor-critic manner to a multi-agent system. \cite{neurips17_maddpg} proposes a multi-agent framework that uses DDPG \cite{iclr16_ddpg} to update the decentralized policy and a centralized critic, and some works \cite{icml19_maac,neurips18_atoc} further integrate the critic network's attention mechanism to avoid the critic biased by irrelevant information. 
        
In some scenarios, an agent can't get other agents' local observation \cite{icml20_smp,neurips19_self_assembly}. In this case, communication is the need, so the agent can exchange information to achieve cooperative or competitive behavior \cite{neurips16_commnet}. Some works \cite{neurips18_atoc,icml19_tarmac,aamas21_magic,aaai20_ga_comm,aamas21_dicg} compose the attention mechanism to improve communication efficiency, while others \cite{iclr19_ic3net} design gating mechanism to decide when to communicate.

\textcolor{neurips}{
	In addition, the multi-agent environment could be non-stationary since the agents are changing at the same time. Some works \cite{iclr18_contadapt} view a non-stationary task as sequential stationary tasks and apply MAML to achieve fast adaptation while MetaMAPG and DiCE \cite{arxiv21_metamapg,icml18_dice} proposed a meta-learning gradient theory to accounts for the non-stationary setting. 
	However, these works assume each agent is controlled by its own policy, and all the policies are not shared. This kind of problem formulation rises the potential difficulty of adaptation for different number of agents, and they don't include the experiments about adapting to different number of agents.
}
\textcolor{aaai}{
Some works \cite{iclr20_epc,aaai20_few2more} make use of curriculum learning to continually adapt to large number of agents in the environment.}
\section{Method}
\begin{figure*}[t]
	\centering
	\includegraphics[width=0.95\textwidth]{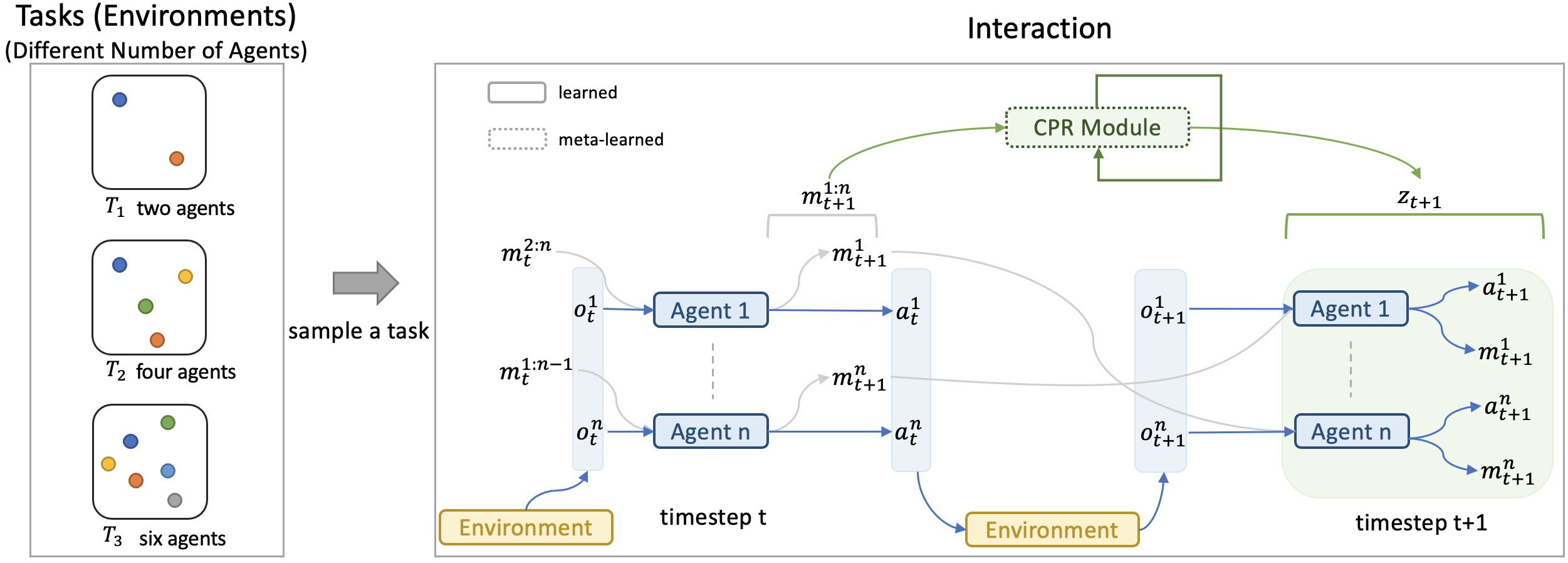}
	\caption{
		The overview of the proposed framework. The framework first samples a task $T$ (or an environment), then interact with the environment to get the a series of transitions $(o, a, r)$. During the interaction, the agent takes as inputs (a) its local observation and (b) messages from other agents, and (c) the task-specific information from the CPR module; then the agent predicts the action distribution and forms its message to be delivered to other agents. The CPR module distills information from the history of communication among agents to predict the task-specific information $z$.
	}
	\label{fig:overview}
	\vspace{-8pt}
\end{figure*}

We first provide the problem statement in Sec. \ref{sec2:problem_statement}. Then, we describe the architecture of the proposed framework in Sec. \ref{sec2:architecture}, followed by the training workflow in Sec. \ref{sec2:training}.
\subsection{Problem Statement} \label{sec2:problem_statement}
We formulate a multi-agent task as a Partially Observable Markov Decision Process (POMDP) \cite{neurips17_maddpg}. Specifically, a Markov game for $n$ agents is deﬁned by a set of states $\mathbf{S}$ describing the possible conﬁgurations of all agents, a set of actions $\mathbf{A}_1, ..., \mathbf{A}_n$ and a set of private observations $\mathbf{O}_1, ..., \mathbf{O}_n$ for each agent. Therefore, a multi-agent task is described as $\mathbf{T} = \{\mathbf{S}, \mathbf{A}_{1:n}, \mathbf{O}_{1:n}, n\}$.
        
Each agent $i$ uses a stochastic policy $\pi_{i}: \mathbf{O}_i \times \mathbf{A}_i \longrightarrow [0, 1]$ to select an action, which leads to the next state according to the transition function $B: \mathbf{S} \times \mathbf{A}_1 \times ... \times \mathbf{A}_n \longrightarrow \mathbf{S}$. Then, each agent $i$ gets rewards as a function of the state and agent’s action $r_i: \mathbf{S} \times \mathbf{A}_i \longrightarrow \mathbf{R}$, and receives an observation correlated with the state $o_i: \mathbf{S} \longrightarrow \mathbf{O}_i$ . The initial states are defined by a distribution $\sigma:\mathbf{S} \longrightarrow [0, 1]$. Each agent $i$ intends to maximize their own total expected return $R_i = \sum_{t=1}^{L} \gamma^t r _t^i$ where $\gamma$ is a discount factor and $L$ stands for the maximum of steps an agent can take.
        
In this work, we first train the MARL framework in multi-agent tasks ${\mathbb{T}}_{train} = \{\{\mathbf{S}, \mathbf{A}_{1:n}, \mathbf{O}_{1:n}, n\}, n \in \mathbf{N}_{train}\}$. The idea is to make the framework be able to quickly adapt to 
\textcolor{neurips}{
	${\mathbb{T}}_{adapt} = \{\{\mathbf{S}, \mathbf{A}_{1:n}, \mathbf{O}_{1:n}, n\}, n \in \mathbf{N}_{adapt}\}$. 
}
Note that the number of agents in the adaptation phase later will always be larger than the number of agents in the training phase, i.e., 
\begin{equation}
	a > b,\; \forall a \in \mathbf{N}_{adapt}, \forall b \in \mathbf{N}_{train}
\end{equation}

\subsection{Proposed Framework} \label{sec2:architecture} 
\begin{figure}
	\centering

	\includegraphics[width=0.47\textwidth]{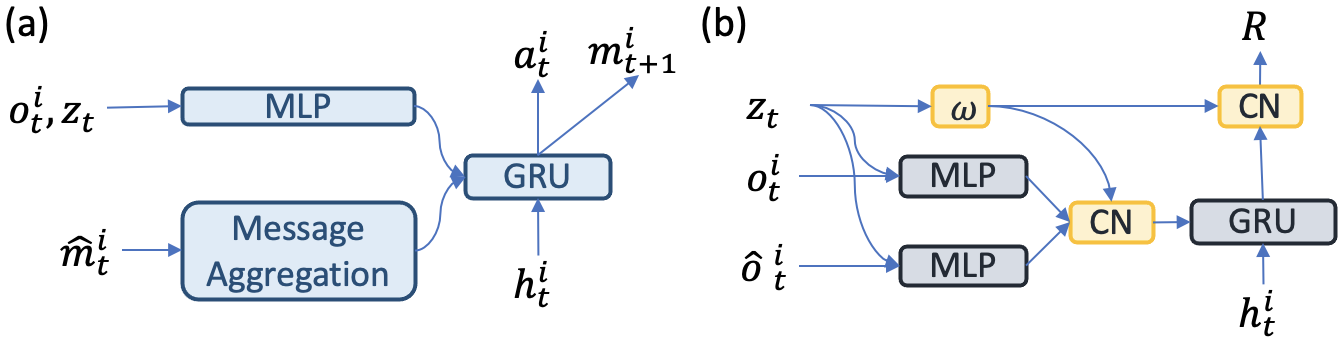}
	\caption{
		(a) The policy network architecture. The policy network takes the local observation $o$, task-speciﬁc information $z$ and messages $\hat{m}$ from other agents as input. Then, the policy network predicts action distribution $a$ and message $m$ based on its hidden state $h$ (b) The critic network architecture. The critic network predicts an agent's average return based on its local observation $o$, the observation $\hat{o}$ from other agents and the task-speciﬁc information $z$ and its hidden state $h$. 
	}
	\label{fig:ac_arch}
	\vspace{-8pt}
\end{figure}
        
We provide the details of our multi-agent communication framework shown in Fig. \ref{fig:overview} to solve the problem defined in Sec. \ref{sec2:problem_statement}. In an environment, we have $n$ agents which are controlled by a shared stochastic policy $\pi_\theta$ that contain recurrent hidden state $h$ parameterized by $\theta$. Since $i$-th agent only accesses local observation $o_t^i$ in partial-observable environment, it sends $m_{t}^i$ and receives messages $\hat{m}_{t}^{i}= \{m_{t}^j \,|\, j \in \{1,2,...,n\}, j\neq i \}$ sent from other agents to form cooperative behavior at timestep $t$ (see Sec. \ref{sec2:communication_mechanism} for further illustration). A task-specific information $z_{t}$ shared across agents further guides the agents to perform well in a task $\mathbf{T}$ (see Sec. \ref{sec:cpr} for further illustration). Therefore, the action $a_t^i$ of $i$-th agent is sampled from the policy, i.e. $a_t^i \sim \pi_{\theta}(a_t^i | o_t^i, h_{t}^i, \hat{m}_{t}^{i}, z_{t})$.
A critic network $V_\mu(o_t^i, \hat{o}_t^{i}, h_{t}^i, z_t)$ that contains recurrent hidden state $h_t^i$ followed by Conditional Normalization (CN) \cite{aaai18_film} predicts the returns of $i$-th agent, where $\hat{o}_t^{i}$ represents the local observations of all agent except agent $i$. A CN weight generator $\omega$ takes task-specific information $z_t$ as input and generates normalization offset and scale, so the CN weight generator $\omega$ learns to normalize the average return according to $z_t$. For detailed information of CN, please refer to \cite{aaai18_film}.


\subsubsection{Communication Mechanism} \label{sec2:communication_mechanism}
\begin{figure}
	\centering
	\includegraphics[width=0.43\textwidth]{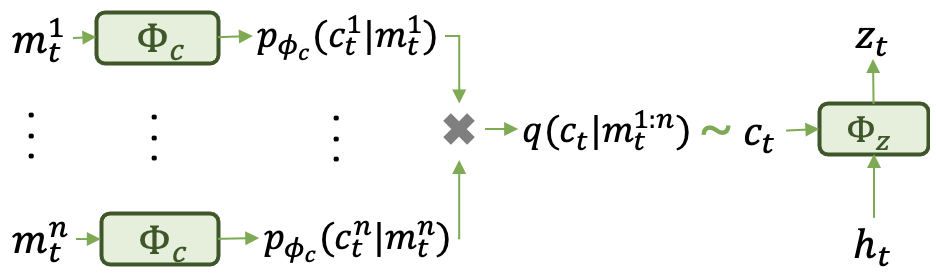}
	\caption{The architecture of CPR module. At each timestep, the message $m$ of each agent are fed in to a latent context estimator $\Phi_c$ to estimate a latent distribution of a single message. Then, the context distribution of each message are fused and passed to GRU to predict the task-specific information $z$.}
	\label{fig:cpr}
	\vspace{-8pt}
\end{figure}
The communication mechanism in our framework can be formed into two steps. In the first step, the received messages are broadcast to all the agents, and the agent encodes the received messages by message encoder $F$. An average operation fuses the encoded messages:
$e_t^i = \frac{1}{n-1} \sum_{j=1, j \neq i}^n F(m_t^i)$.
In the second step, the local observation will be encoded by observation encoder $G$. Then, the encoded observation and aggregated messages are concatenated and fed to the GRU \cite{emnlp14_gru} to form $h_{t+1}^i = GRU(h_{t}^i, e_t^i, G(o_t^i))$.

Finally, the output of the GRU is used to predict the action distribution as well as message for the next timestep.
        
\subsubsection{CPR Module} \label{sec:cpr}
\textcolor{aaai}{
Human can briefly understand a thing via listening to the communication within a group of people, and we believe multi-agent also has this property. Thus, we propose CPR module that aims at summarizing the communication history and extracting the difference between the tasks since we believe successful communication behaviors should contain compact information for a task. 
The architecture of the CPR module is presented in Fig. \ref{fig:cpr}.
}
The CPR module contains a latent context estimator $\Phi_c$ parameterized by $\phi_c$ that compresses the information in the message into a latent distribution $p_{\phi_c}(c|m)$ and a recurrent context encoder $\Phi_z$ parameterized by $\phi_z$ which contains recurrent hidden state $h_t$ to comprise contexts so far to form task-specific information $z_t$. We form $\phi = [\phi_c, \phi_z]$ as the parameters of the whole CPR module. 
\textcolor{aaai}{The reason to use probabilistic estimation for task distribution is that it can potentially cover larger task distribution and improve the transferability between different tasks.}
To be more specific, at each timestep, the message of each agent is modeled into Gaussian latent distribution $p_{\phi_c}(c_t^i|m_t^i)$ by variational inference. Then, the latent distribution of each message is fused by the multiplication of probability distribution to form a shared context distribution $q(c_t|m_t^{1:n})$ across all the agents. A context variable $c$ is sampled from $q(c_t^i|m_t^i)$ and fed to recurrent context encoder $\Phi_z$ to predict a task-specific vector $z_t$ which is shared across all the agents: 
\begin{equation}
	\label{eq:cpr}
	z_t = \Phi_z(c_t, h_{t}), \;\; c_t \sim q(c_t|m_t^{1:n})
\end{equation}
 
\begin{algorithm}[tb]
	\caption{Meta-CPR Training}
	\label{alg:meta_cpr}
	\begin{algorithmic}
		\STATE {\bfseries Required:} A set of multi-agent task ${\mathbb{T}}_{train}$, 
		policy network $\theta$, critic network $\mu$, CPR module $\phi$ 
		information bottleneck weight $\lambda$, 
		learning rates $\alpha_1$, $\alpha_2$, $\alpha_3$
		\WHILE{not done}
		\STATE// Collect trajectory
		\FOR{each $\mathbf{T}_i$ in ${\mathbb{T}}_{train}$}
		\STATE Initialize buffer $D_i$ to store trajectories
		\STATE Generate $k$ episodes with $\pi_\theta(a|\cdot)$ and add it to $D_i$
		\ENDFOR
		\STATE// Calculate losses
		\FOR{each $\mathbf{T}_i$ in ${\mathbb{T}}_{train}$}   
		\STATE $L_{P}^i = L_{P}(D_i)$;$L_{C}^i = L_{C}(D_i)$;$L_{KL}^i = L_{KL}(D_i)$
		\ENDFOR
		\STATE// Update the framework
		\STATE $\theta \leftarrow \theta + \alpha_1 \nabla_\theta \sum_i L_{P}^i$
		\STATE $\mu \leftarrow \mu - \alpha_2 \nabla_\mu \sum_i L_{C}^i$
		\STATE $\phi \leftarrow \phi - \alpha_3 \nabla_\phi \sum_i (L_{KL}^i+L_{C}^i)$
		\ENDWHILE
	\end{algorithmic}
\end{algorithm}
\vspace{-8pt}

\subsection{Training} \label{sec2:training}
\begin{figure}
	\centering
	\includegraphics[width=0.43\textwidth]{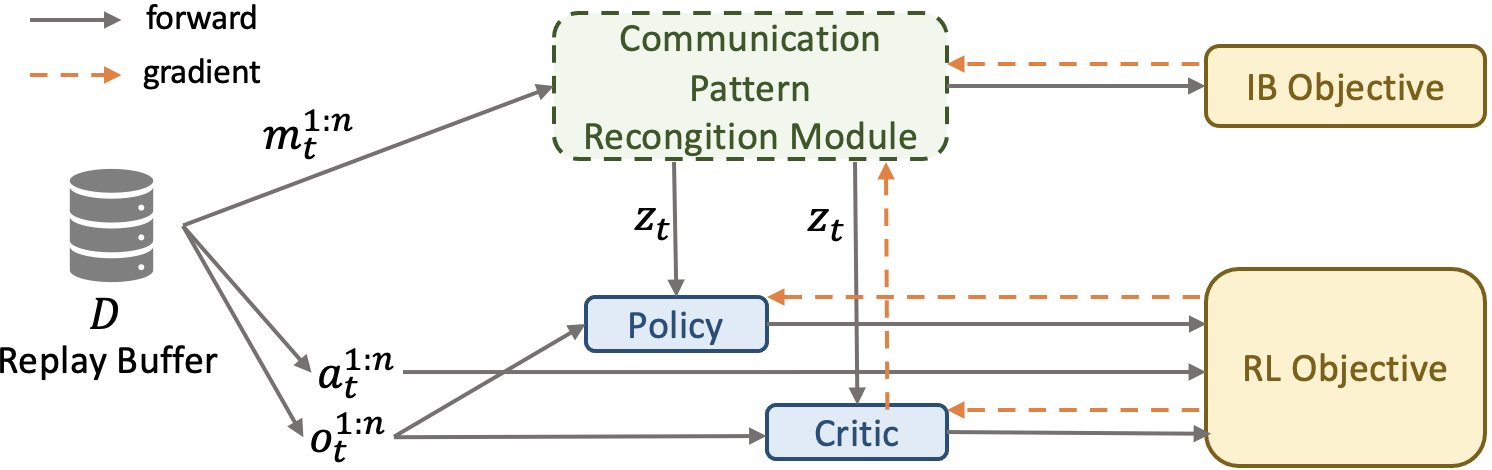}
	\caption{The training pipeline of our method. Both policy and critic are updated with A2C based on the observation $o$ , action $a$ and the messages $m$ in replay buffer $D$, and the CPR module is jointly updated with the critic as well as the Information Bottleneck objective.}
	\label{fig:training}
	\vspace{-8pt}
\end{figure}
The whole policy network and critic network are optimized with A2C \cite{icml16_a2c}. The return $(R)$ is estimated with GAE \cite{iclr16_gae}. A replay buffer $D$ is used to store the interaction trajectories. The training pipeline is shown in Fig. \ref{fig:training}.
\textcolor{neurips}{We hypothesize that the reward scale is the most obvious signal to capture the task distribution, so we only meta-learn the CPR module with the gradient from the critic. 
}
Hence, the extracted task $z_t$ tends to become a kind of representation that can enhance the return of $o_t$. The CPR module is also meta-learned by the Information Bottleneck (IB) \cite{iclr16_vib} weighted by $\lambda$. Therefore, the $z_t$ tend to contain the noiseless information and preserve information that aims at maximizing average return. In other words, the CPR module learns how to provide meaningful information for downstream policy to learn to solve various MDPs corresponding to different numbers of agents. 
       
\begin{figure*}[t]
	\centering
	\includegraphics[width=0.95\textwidth]{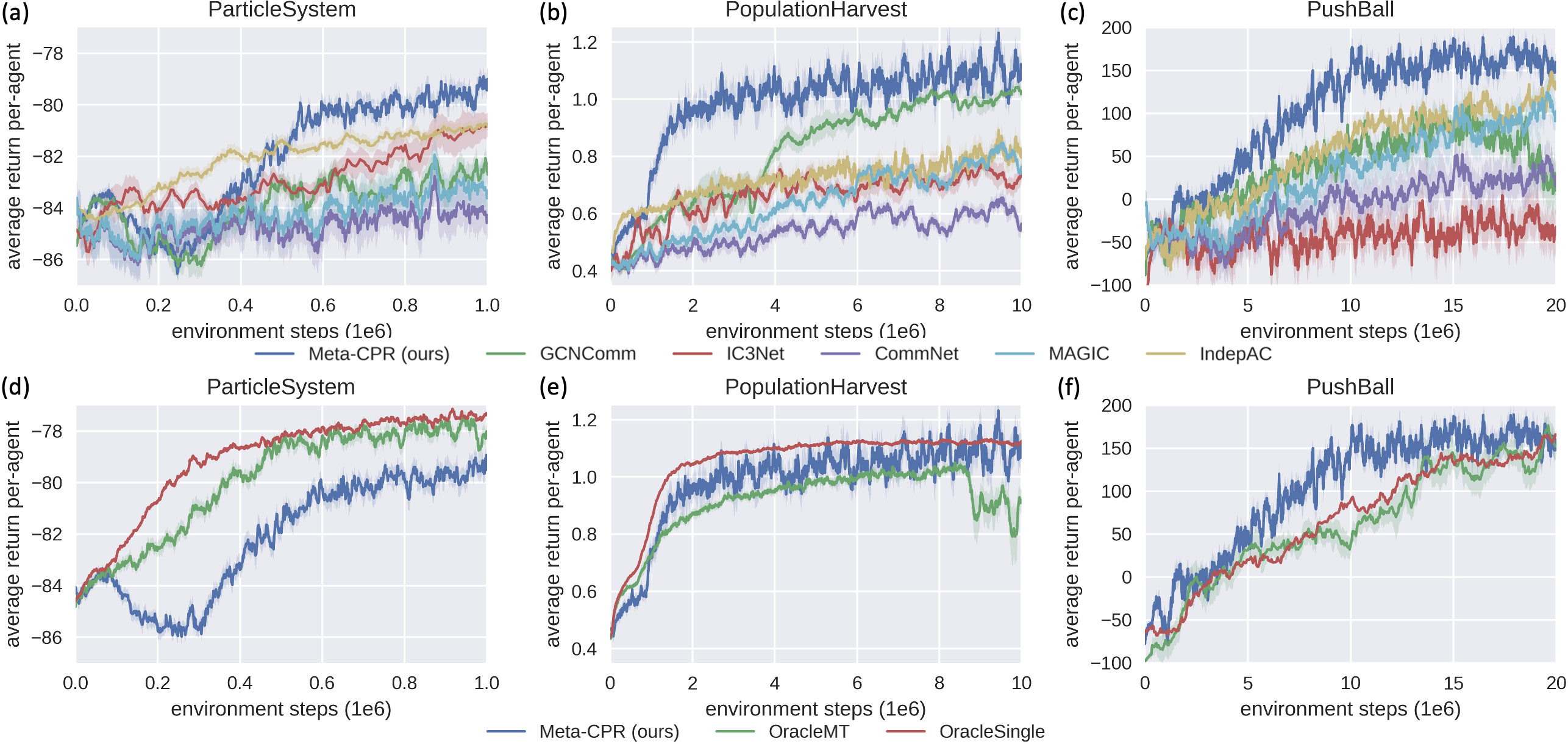}
	\caption{
		(a-c) Zero-shot generalization performance to the unseen large number of agents. 
		Our method outperforms other baselines.
		Each learning curve is merged with ten random seeds, and the shaded area represents the standard error. 
		(d-f) Compared with the oracle method in zero-shot transfer performance to the unseen large number of agents. 
		The oracle methods outperform our method since they directly fit in the environment with a large number of agents in the easy scenario such as ParticleSystem, but our method reach comparing and better performance in PopulationHarvest and PushBall respectively.
	}
	\label{fig:main_performance}
\end{figure*}  
        
To be more specific, the objective for training the policy $L_P$ is shown below,
\begin{equation}
	L_{P} = E_D[\log \pi_\theta(a|o,h,m,z) (R - V_\mu(\cdot)) + \eta H(\pi_\theta(\cdot))]  
\end{equation}
where $H(\pi(\cdot))$ represents the entropy of the distribution and is used to encourage the exploration.
Note that the gradient of $L_{P}$ is not propagated to the CPR module. The critic objective $L_C$ is similar to normal A2C objective, which is $L_{C} = E_D[(V_\mu(o, \hat{o}, h, z)-R)^2]$.

The gradient of $L_{C}$ is propagated to the CPR module, so the CPR won't extract task-irrelevant information for the task. Finally, the KL divergence is applied to the CPR module as IB,
i.e. $L_{KL} = \lambda \cdot D_{KL}(p_{\phi_c} (c|m) || p(c))$,
where $p(c)$ is unit Gaussian prior over $c$. Note that the messages used to update the CPR module is derived from the replay buffer $D$ that contains the past message used in the previous interaction. The whole algorithm is summarized to Algorithm. \ref{alg:meta_cpr}.


\begin{figure}
	\begin{center}
		\includegraphics[width=0.47\textwidth]{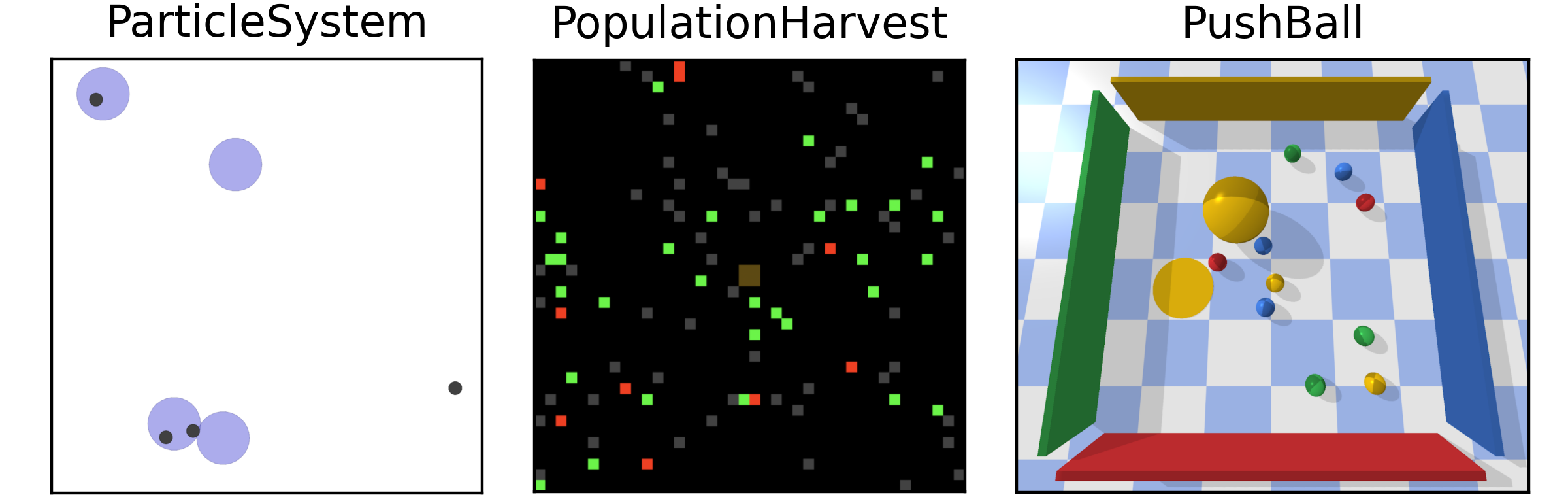}
	\end{center}
	\caption{Environments used in our experiments. All of the environments belong to the fully cooperative scenario. See Sec. \ref{sec4:environment} for more information.
	}
	\label{fig:env}
	\vspace{-12pt}
\end{figure}

\section{Experiments}
    
We first describe the environmental setting including baselines and multi-agent environments in Sec. \ref{sec4:environment}. Then, the main results are demonstrated in Sec. \ref{sec4:main_result}. The ablation study is shown in Sec. \ref{sec4:ablation_study}. We further visualize the behavior of the multi-agent framework in Sec. \ref{sec4:vis}.
   
\begin{figure*}[t]
	\centering
	\includegraphics[width=0.95\textwidth]{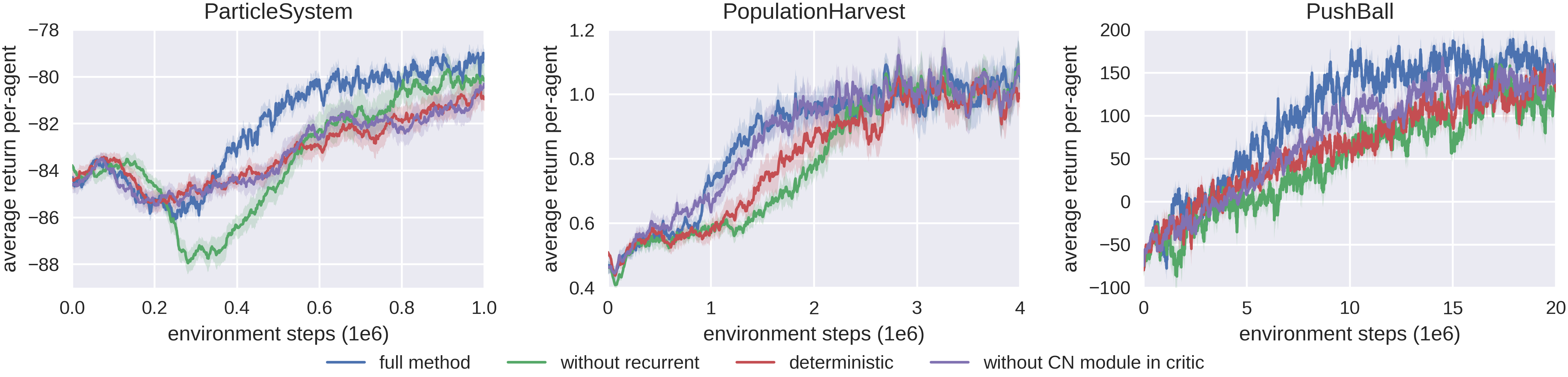}
	\caption{Ablation study on our method. The zero-shot transfer learning curves show that our full method outperforms other ablation designs. Each learning curve is merged with ten random seeds, and the shaded area represents the standard error.}
	\label{fig:ablation_study}
\end{figure*}   
\begin{figure*}
	\centering
	\includegraphics[width=0.95\textwidth]{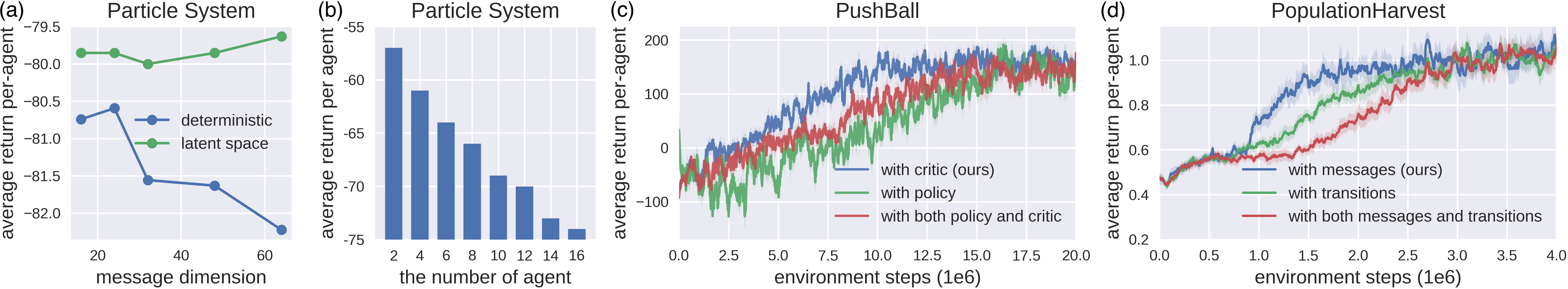}
	\caption{
		(a) The average return of GCNComm in zero-shot transfer scenarios when the message dimension varies. The performance of the deterministic version drops when the message dimension increases.
		(b) The average return of each agent of training on a specific number of the agent. We find that the scales of the performance are different.
		\textcolor{neurips}{
    		(c) The zero-shot transfer learning curves of jointly training the CPR module in the different manners. 
    		(d) The zero-shot transfer results of using different information as the context. 
		}
	}
	\label{fig:cond_norm_ib}
	\label{fig:encoder_training}
	\label{fig:context}
	\vspace{-8pt}
\end{figure*}  
   
\subsection{Environmental Setup} \label{sec4:baseline} \label{sec4:environment}
We compare our method with other baselines in three environments (see Fig. \ref{fig:env}). \textbf{ParticleSystem}: An environment adapted from \cite{arxiv_language_ma}. Each agent (blue) needs to reach a landmark (black) given the landmark position, and the agent's observation doesn't contain the other agents' positions. During the agent's movement, the agents try to prevent the collision, so communication is essential. \textbf{PopulationHarvest}: The agents (red) intend to carry an apple (green) to the target position (brown). The agents need to avoid trying to get the same apple. \textbf{PushBall}: The agents need to cooperate to push a heavy ball to a target position. Note that the ball's weight is larger than that of the agent, so cooperative behavior is required. The environment is constructed with PyBullet \cite{pybullet}. Note that both ParticleSystem and PopulationHarvest are discrete action space, and PushBall is the continuous one.
    
\textcolor{aaai}{
Our method is compared with multi-agent communication frameworks, i.e. IC3Net ~\cite{iclr19_ic3net}, CommNet~\cite{neurips16_commnet} and MAGIC~\cite{aamas21_magic}. We also design some baselines as shown below to illustrate the properties of our approach.
}
%
\textbf{GCNComm}: Our method without using the CPR module, and this is effectively equivalent to use Graph Convolution Network (GCN) \cite{iclr17_gcn} as communication mechanism. This baseline can help us to understand the effectiveness of the CPR module clearly. \textbf{IndepAC}: We use a naive actor-critic architecture to train the policy without communication. It is helpful for us to understand the difference between the multi-agent framework as well as single-agent training. \textbf{OracleMT}: We directly train GCNComm on the environments with the number of agents used in the adaptation phase. \textbf{OracleSingle}: We directly train GCNComm on the environments with for the specific number of agents used in the adaptation phase. \textcolor{neurips}{We also provide more baselines and comparison in the supplementary Sec 1.
The whole framework is update by Adam optimizer \cite{iclr15_adam}, and the detailed hyperparameter settings are shown in the supplementary Sec 3.}
    
\subsection{Experimental Result} \label{sec4:main_result}
The multi-agent frameworks are trained with the environments which contain $n$ agents, where $n \in \mathbf{N}_{train}$. Then, we make the frameworks perform the zero-shot transfer to environments that contain $b$ agents, where $b \in \mathbf{N}_{adapt}$. In Fig. \ref{fig:main_performance}, we present the average return per-agent tested on environments with the unseen number of agents. For the detailed experiment setting, please refer to the supplementary.
        
From the comparison between Meta-CPR and GCNComm, we find that the CPR module helps the agent zero-shot adapt to the environment with a larger number of agents. 
As far as IC3Net, CommNet and MAGIC, IC3Net outperforms CommNet and MAGIC in the sample efficiency. We hypothesize that the communication gating mechanism makes communication more robust when the number of agents changes. The gating mechanism makes agents not over-rely on communication. 
However, our method still outperforms IC3Net. To our surprise, IndepAC beats some multi-agent framework in the adaptation performance. Note that the multi-agent framework outperform IndepAC in the training phase. We hypothesis that it is because the IndepAC makes action distribution solely based on its private observation. As a consequence, though the number of agents becomes larger and the reward distribution change accordingly, IndepAC still performs relatively reasonable behavior. In contrast, the multi-agent framework predicts action distribution partially based on the communication correlated to the number of agents.
        
As for the comparison between our method and the oracles (Fig. \ref{fig:main_performance}), our approach reaches comparing performance compared with the oracles in PopulationHarvest and PushBall. We hypothesis that since the environment with a large number of agents is more complex than that with a small one, directly train the framework on lots of agents may suffer from low sample efﬁciency.
This also implies that learning with the small number of agents and transferring to a large number of agents is potentially more efﬁcient than directly training with a large number of the agent in some cases.

\begin{figure*}[t]
	\centering
	\includegraphics[width=0.95\textwidth]{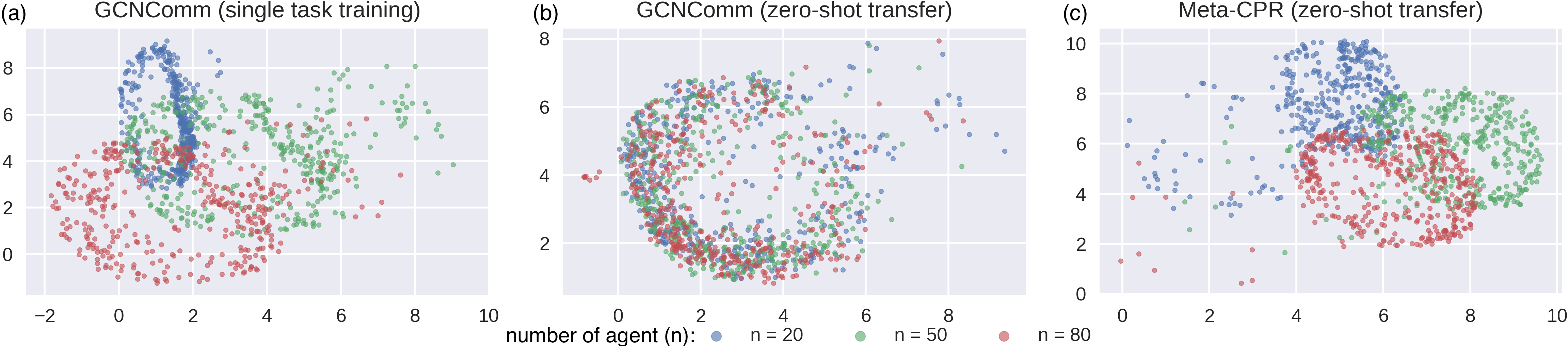}
	\caption{Embedding of the trajectories generated by different multi-agent frameworks in PopulationHarvest. (a) Directly training GCNComm, i.e. OracleSingle described in Sec. \ref{sec4:baseline}, on a specific number of agents does induce different behavior.  (b) Using GCNComm to perform zero-shot transfer can't fit the tasks with the different number of agents. (c) In contrast, zero-shot transfer with our method indeed performs different behavior for a specific number of agents.}
	\label{fig:trajectory_embedding}
    \vspace{-8pt}
\end{figure*}  

\subsection{Ablation Study} \label{sec4:ablation_study}
We conduct ablation study including framework architecture variants, different framework strategies as well as context modeling strategies.


\subsubsection{Framework Architecture}
We perform an ablation study to explore
1) the necessity of using recurrence in the CPR module, 2) the benefit of using probabilistic estimation in the CPR module, 3) the advantage of leveraging CN module in the critic. The results are presented in Fig. \ref{fig:ablation_study}.
We clearly find that our full method outperforms other ablation designs. The CPR module without recurrent module performs the worst since the CPR module can't derive temporal information during the interaction. Moreover, The variant of the CPR module without probabilistic estimation is a deterministic version of the CPR module by reducing the distribution $q(z|m)$ to a point estimate. From the comparison between our method and the deterministic CPR module, our method can achieve greater zero-shot transfer performance. We hypotheses that it is because the probabilistic estimation potentially covers larger task distribution and improves the transferability between different tasks.
We also observe that the deterministic approach's performance gets worse as the message dimension increases (Fig. \ref{fig:cond_norm_ib} (a)), which implies probabilistic estimation is potentially more robust to the input dimension.
Besides, the returns in the environments with different number of agents local in different scale from observing results of training GCNComm on the environment with only one kind of amount of agents as shown in Fig. \ref{fig:cond_norm_ib} (b). The return distribution shifting may cause the performance drop in the zero-shot transfer scenario. The CN module can mitigate this issue, and more details are shown in supplementary.
            
\subsubsection{Training Strategies}

The CPR module should be jointly trained with RL objective, so it can infer what kinds of information is useful for the training process. We jointly train the CPR module with the critic, and we further compare it with other training methods. That is 1) jointly training with both critic and policy network 2) jointly training with only the policy network.
\textcolor{neurips}{
From Fig. \ref{fig:encoder_training} (c), we find that jointly training CPR module with the only critic obtains better performance, which supports the hypothesis that the reward scale is the most obvious signal to capture the task distribution.
}
In contrast, the policy performs similar behavior across different tasks at the beginning of training, so it is relatively hard for the CPR module to understand the difference between the tasks from the policy. Hence, jointly training CPR module with only the policy is the least efficient one.


\subsubsection{Context Modeling Strategies}
            
Recall that the CPR module tries to derive task-specific information to benefit the training and generalization to the unseen large number of agents. Since messages are lightweight information to facilitate cooperative behavior between the agents, we use messages as the input of the CPR module. However, we can instead leverage 1) the environment transitions $(o,a,t)$ 2) both transitions and messages to infer task-specific information.
\textcolor{neurips}{Fig. \ref{fig:context} (d) demonstrates that using the only message as the context is the most efficient approach, and it meets our belief that the messages are likely already to contain refined and vital information of its transition. }
As a result, observing the full transitions won't contribute more helpful information to the agent in the multi-agent scenario, and the CPR module may need to use more training steps to filter out redundant information. It is also consistent that using both message and transition is less efficient.

\subsection{Visualization of Multi-Agent Framework Behavior}
\label{sec4:vis}
To further demonstrate the difference of the multi-agent frameworks, we visualize the embedding of the trajectories generated with different methods with different numbers of agents. To be more specific, we first concatenate the transitions $(o_t, a_t, r_t)$ of a trajectory into a vector,  and we use UMAP \cite{arxiv20_umap} to project the high-dimensional trajectory information to 2D embedding space. In Fig. \ref{fig:trajectory_embedding} (a), the GCNComm trained in the environment with a different number of agents does express different behavior. From observing the episodes during the learning process, we find that the agent tries to get an object in a greedy manner if the number of agents is small. However, when the number of agents becomes larger, some agents try to get objects while other agents tend to explore to find other objects. Intuitively, if a multi-agent framework can be successfully adapted to different number of agents, it should successfully perform different behavior according to the number of agent in the environment. In Fig. \ref{fig:trajectory_embedding} (b), the zero-shot transfer result of GCNComm in the environment with different number of agents performs similar behavior, and it implies that GCNComm can't performs suitable behavior according to the number of agents. In contrast, the zero-shot transfer result of our method in the environment with different number of agents performs different behavior shown in Fig. \ref{fig:trajectory_embedding} (c), and it suggest that our method can successfully fit to the specific tasks in zero-shot transfer scenarios. 
\textcolor{aaai}{We also describe how the behavior of the agent changes as the number of agents varies in supplementary Sec 5.}
\section{Conclusion}
We propose Meta-CPR, a multi-agent framework that can be generalized to a large unseen number of agents. Our method reformulates the multi-agent task with a different number of agents as a multi-tasking problem. Then we learn a module to capture the communication pattern in a context-based meta-learning fashion. We show that our method outperforms other multi-agent frameworks in the adaptation performance in both discrete and continuous action space environments. Moreover, we further conduct ablation studies to show the Meta-CPR variations of the module's design, the various context options, and the training strategy. We hope this work can inspire researchers that focus on applying MARL frameworks the real-world environments to tackle the generalization between the different number of agents.

\bibliography{reference}

\end{document}